\documentclass{article}
\usepackage{ijcai24}

\usepackage{array}
\usepackage{times}
\usepackage{soul}
\usepackage{url}
\usepackage[hidelinks]{hyperref}
\usepackage[utf8]{inputenc}
\usepackage[small]{caption}
\usepackage{graphicx}
\usepackage{booktabs}
\usepackage[switch]{lineno}
\usepackage{subfigure}
\usepackage{booktabs}
\usepackage{bbold}
\usepackage{multirow}
\usepackage[ruled,vlined,linesnumbered]{algorithm2e}
\usepackage{enumerate}
\usepackage{amsmath,amsthm,amssymb,amsfonts}
\usepackage{bm}
\usepackage{bbm}
\usepackage{color}

\title{Estimating before Debiasing: A Bayesian Approach to Detaching Prior Bias in Federated Semi-Supervised Learning}

\author{
Guogang Zhu$^1$
\and
Xuefeng Liu$^{1,2}$\and
Xinghao Wu$^1$\and
Shaojie Tang$^3$\and
Chao Tang$^1$\and \\
Jianwei Niu$^{1,2}$\footnote{Jianwei Niu is the corresponding author.}\And
Hao Su$^1$
\affiliations
$^1$State Key Laboratory of Virtual Reality Technology and Systems, School of Computer Science and Engineering, Beihang University, Beijing, China \\
$^2$ Zhongguancun Laboratory, Beijing, China \\
$^3$Jindal School of Management, The University of
Texas at Dallas, Richardson, TX, USA\\
\emails
\{buaa\_zgg, liu\_xuefeng, wuxinghao\}@buaa.edu.cn,
shaojie.tang@utdallas.edu,
\{sy2106322, niujianwei, bhsuhao\}@buaa.edu.cn
}

\begin{document}

\maketitle

\begin{abstract}
Federated Semi-Supervised Learning (FSSL) leverages both labeled and unlabeled data on clients to collaboratively train a model. 
In FSSL, the heterogeneous data can introduce prediction bias into the model, causing the model's prediction to skew towards some certain classes. 
Existing FSSL methods primarily tackle this issue by enhancing consistency in model parameters or outputs. 
However, as the models themselves are biased, merely constraining their consistency is not sufficient to alleviate prediction bias. 
In this paper, we explore this bias from a Bayesian perspective and demonstrate that it principally originates from label prior bias within the training data.
Building upon this insight, we propose a debiasing method for FSSL named FedDB. FedDB utilizes the Average Prediction Probability of Unlabeled Data (APP-U) to approximate the biased prior.
During local training, FedDB employs APP-U to refine pseudo-labeling through Bayes' theorem, thereby significantly reducing the label prior bias. 
Concurrently, during the model aggregation, FedDB uses APP-U from participating clients to formulate unbiased aggregate weights, thereby effectively diminishing bias in the global model. 
Experimental results show that FedDB can surpass existing FSSL methods.
The code is available at \href{https://github.com/GuogangZhu/FedDB}{https://github.com/GuogangZhu/FedDB}.
\end{abstract}

\section{Introduction}

Federated Learning (FL) \cite{FedAvg} is a distributed learning paradigm that can facilitate collaborative model training among multiple clients while preserving data privacy. 
Presently, most FL methods are confined to supervised learning (SL) settings, wherein it is presumed that each client maintains a fully labeled dataset.
Nevertheless, in real-world applications, data labeling is notably laborious and time-consuming.
Therefore, a more realistic case involves each client possessing a mix of unlabeled and labeled data.
This specific scenario, known as Federated Semi-Supervised Learning (FSSL), has been explored in various studies \cite{FedMatch,lin2021semifed,SemiFL} and is garnering increasing interest within the FL research community. 

In this study, we focus on an FSSL setting where the data on each client are class-imbalanced.
Moreover, it is assumed that both intra-client and inter-client data heterogeneity exist.
Specifically, intra-client data heterogeneity implies that both the labeled data and unlabeled data on an individual client originate from diverse distributions.
Inter-client data heterogeneity means that the overall distributions across clients are non-independent and identically distributed (Non-IID). 

\begin{figure}
    \centering
    \includegraphics[width=3.3in]{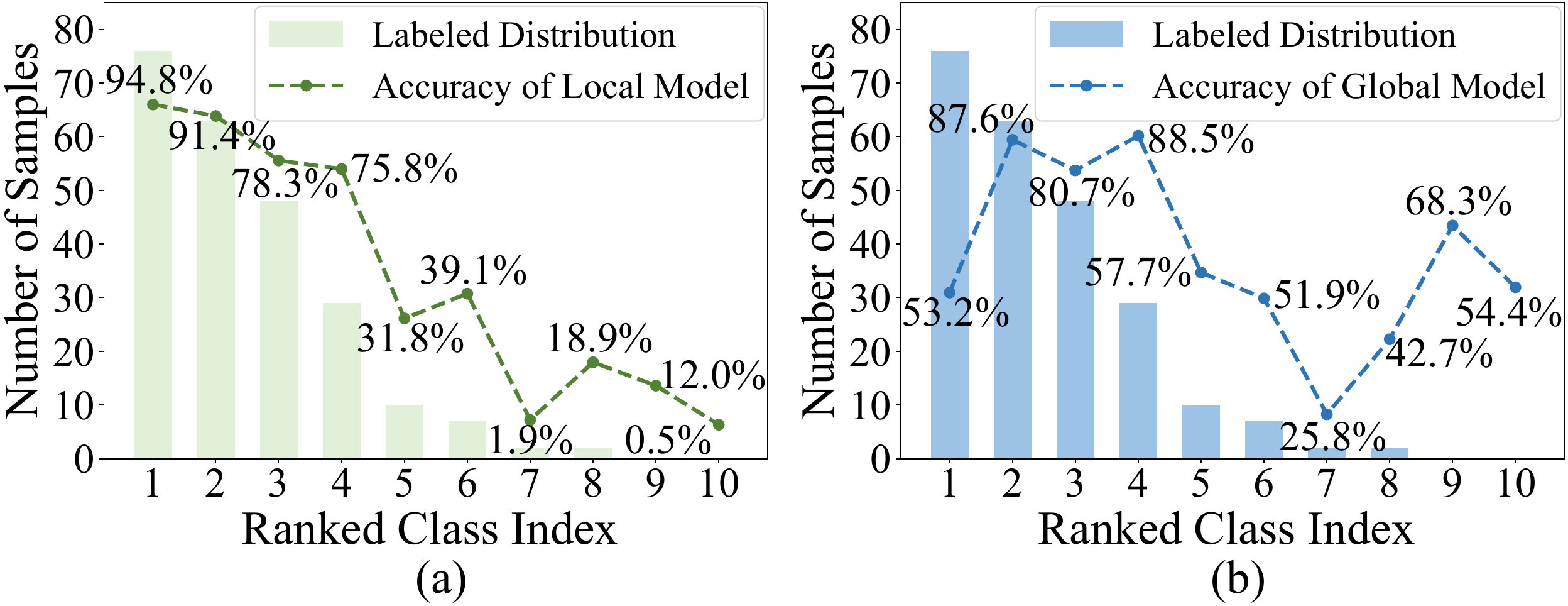}
    \caption{Class-wise test accuracy on a balanced test dataset, along with the labeled data distribution on an individual client. (a) Test accuracy of local model, (b) Test accuracy of global model. The class indexes are ranked based on the labeled data distribution.}
    \label{class-wise-accuracy}
\end{figure}

In the described scenario, the model's prediction can skew to some certain classes during the training, i.e., prediction bias.
Figure \ref{class-wise-accuracy} presents the experimental results conducted in the above scenario, where the overall distributions of labeled and unlabeled data are balanced.
It can be observed that due to class imbalance in the local client, the local model's predictions gradually skew towards the major classes in the local data. 
More importantly, this prediction bias cannot be alleviated after model aggregation, even if the overall distributions are balanced. 
Instead, it evolves into a different form of bias due to the influence from other clients. 
This bias can disrupt the pseudo-labeling process, further creating a `vicious cycle' between pseudo-labeling and local model training.

Existing FSSL methods attribute the above issue to the divergence across clients caused by heterogeneous data and primarily address it by promoting consistency between model parameters or outputs \cite{FedRGD,jiang2022dynamic,liang2022rscfed}. However, as both the local and global models are biased, merely constraining their consistency cannot fundamentally mitigate the model prediction bias.

\begin{figure}
    \centering
   \includegraphics[width=3.3in]{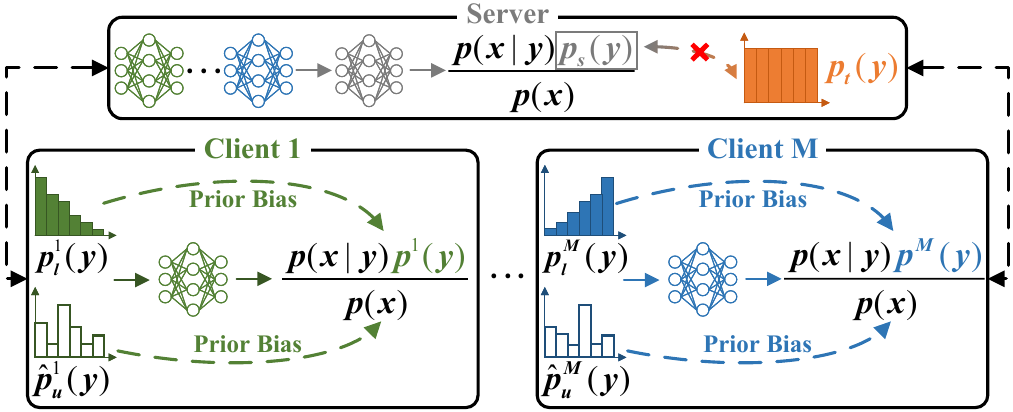}
    \caption{Prior bias in class-imbalanced FSSL.}
    \label{prior-bias}
\end{figure}

In this paper, we delve into the essential reason for the prediction bias in FSSL from a Bayesian perspective. 
Based on Bayes' rule, the model prediction is as follows:
\begin{equation}
    \bm{p}(\bm{y}|\bm{x}) = \frac{\bm{p}(\bm{x}|\bm{y})\bm{p}(\bm{y})}{\bm{p}(\bm{x})},
\end{equation}
where $\bm{p}(\bm{y}|\bm{x})$ is the model's prediction, $\bm{p}(\bm{x}|\bm{y})$ is the class conditional likelihood, $\bm{p}(\bm{y})$ is the label prior.
As shown in Figure \ref{prior-bias}, both the label prior of local labeled data (i.e., $\bm{p}_l(\bm{y})$) and unlabeled data (i.e., $\hat{\bm{p}}_u(\bm{y})$) are biased. Consequently, the model can gradually absorb these biases during training. These biases are eventually injected into the global model through model aggregation, causing its output priors $\bm{p}_s(\bm{y})$ to skew towards certain classes.
When conducting inference on a balanced test dataset (i.e., $\bm{p}_t(\bm{y})$), the model may suffer severe performance degradation, as $\bm{p}_s(\bm{y}) \neq \bm{p}_t(\bm{y})$.

Nevertheless, $\bm{p}_s(\bm{y})$ is commonly challenging to estimate.
On the one hand, in local clients, the ambiguity of pseudo-labels for unlabeled data makes the label prior bias during local training intractable. 
On the other hand, in the server, model aggregation combines influences from participating clients, further complicating the estimation of prior bias.

\begin{figure}[!ht]
    \centering
    \includegraphics[width=3.3in]{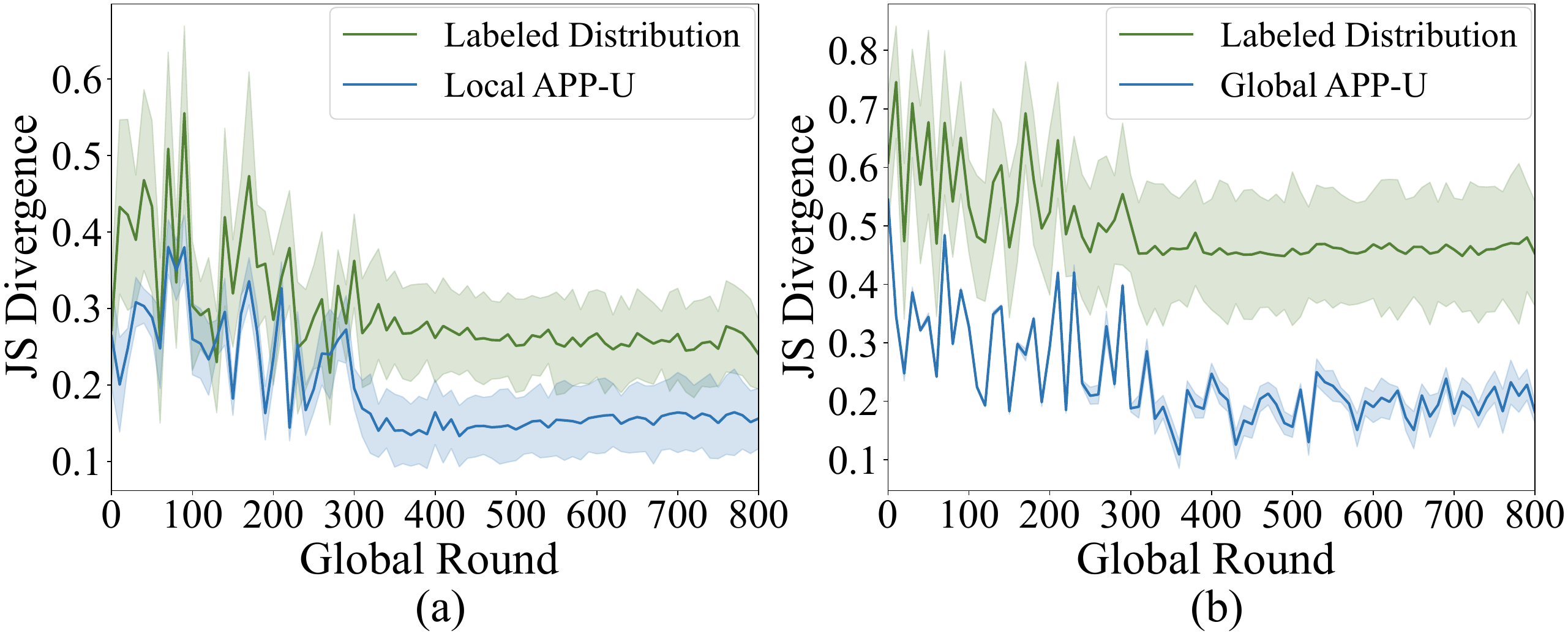}
    \caption{JS divergence between the ground truth bias and either the labeled data distribution or APP-U on clients. (a) Results on the local model, (b) Results on the global model.}
    \label{JS_Divergence}
\end{figure}

Taking the class-wise accuracy on a balanced test dataset as the ground truth for prior bias, we find that the Average Prediction Probability of Unlabeled Data (APP-U) serves as a robust metric to approximate this bias.
Figure \ref{JS_Divergence} illustrates the Jensen–Shannon (JS) divergence \cite{lin1991divergence} between the ground truth bias and either the labeled data distribution or APP-U, where the solid line and shaded area represent the mean and range across clients, respectively.
%for an individual client.
Interestingly, it reveals that for both the global and local models, prior bias does not consistently align with the labeled data distribution. 
Rather, it shows a stronger correlation with APP-U, indicating that APP-U can effectively quantify the prior bias.

Building upon the above insights, we introduce a hierarchical debiasing method for FSSL termed FedDB, to mitigate the prior bias at both the local training and global aggregation stages.  
During the local training, FedDB implements debiased pseudo-labeling (DPL) based on Bayes' theorem, with APP-U serving as the approximation of bias prior. This approach promotes a more balanced pseudo-labeling process for unlabeled data, substantially reducing the label prior bias during local training. 
At the global aggregation stage, FedDB utilizes APP-U from the participating clients to determine optimal aggregation weights.
The above process, termed debiased model aggregation (DMA), effectively mitigates bias within the global model. 
It should be noted that DPL can be seamlessly integrated with FSSL methods that utilize pseudo-labeling with minimal cost. This demonstrates its substantial potential for practical application of FSSL.

The main contributions of this paper are as follows:
\begin{itemize} 
    \item We analyze the prediction bias in class-imbalanced FSSL from a Bayesian perspective.
    \item We propose FedDB, a Bayesian debiasing method for FSSL that uses APP-U as an approximation of prior bias.
    \item We conduct extensive experiments on multiple datasets to demonstrate the effectiveness of FedDB.
\end{itemize}

\section{Related Work}
\subsection{Federated Learning}
Data heterogeneity is a substantial challenge in FL, which can lead to considerable divergence across clients, thereby degrading the model performance \cite{zhao2018federated,li2020federated}. 
To address this issue, various strategies are explored, including reducing the divergence across local models \cite{FedProx,FedDyn,SCAFFOLD}, enhancing aggregation schemes \cite{FedNova,FedDyn,FedOpt}, promoting representation consistency across clients \cite{FedProto,AlignFed,HyperFed}, developing personalized models for individual clients \cite{FedRep,FedDWA}. 
However, these methods primarily focus on SL settings, which is impractical as data labeling is laborious and time-consuming.

\subsection{Semi-Supervised Learning}
SSL aims to mitigate the reliance on labeled data, which prompts various mechanisms to leverage the latent information within unlabeled data.
Pseudo-labeling \cite{lee2013pseudo,wang2023freematch}, also known as self-training, involves assigning pseudo-labels to unlabeled samples with high confidence, enabling their incorporation into the training process.
Consistency regularization \cite{miyato2018virtual} introduces arbitrary perturbations to unlabeled samples and promotes the consistent predictions between different views of unlabeled data.
Additionally, hybrid methods that amalgamate these approaches are also developed, such as MixMatch \cite{berthelot2019mixmatch}, FixMatch \cite{Fixmatch}. 
Recently, SSL has focused on class imbalance, leading to various studies such as class-rebalancing sampling \cite{wei2021crest}, and pseudo label sampling \cite{guo2022class}. However, simply combining these methods with FL is challenging, as they ignore the collaboration across clients.

\subsection{Federated Semi-Supervised Learning}
FSSL can be divided into three distinct scenarios \cite{bai2023combating}: 
(1) \textbf{Labels-at-Partial-Clients}, where
only a few clients have full labels, while the rest possess only unlabeled data \cite{liang2022rscfed,CBAFed}; (2) \textbf{Labels-at-Server}, where labeled data are only available at the server, with local clients merely having unlabeled data \cite{FedRGD,FedMatch,SemiFL}; 
(3) \textbf{Labels-at-Clients}, where each client has mostly unlabeled data and a few labeled samples \cite{FedMatch,bai2023combating}.

This paper focuses on the \textbf{Labels-at-Clients} scenario. 
Currently, several works have been proposed for this scenario.
For instance, SemiFed \cite{lin2021semifed} assigns pseudo-labels to unlabeled data only when multiple models provide consistent predictions. 
FedMatch \cite{FedMatch} enforces prediction consistency across multiple models. 
However, these methods primarily concentrate on encouraging consistency across clients, overlooking the inherent prior biases within the model — a critical factor leading to performance degradation in FSSL with class imbalance.

\section{Preliminary and Background}
In this section, we present the notations used in this paper, followed by a detailed discussion of the framework of FSSL.

\subsection{Problem Setting and Notation of FSSL}
We focus on a FSSL setting for K-class classification task with totally $M$ clients participating in the training.
Each client $m$ maintains a labeled dataset $\mathcal{D}_l^m=\{(\bm{x}^n,\bm{y}^n)\}_{n=1}^{N_l^m}$ and an unlabeled dataset $\mathcal{D}_u^m=\{(\bm{x}^n)\}_{n=1}^{N_u^m}$, where $N_l^m$ and $N_u^m$ are the counts of labeled and unlabeled samples, respectively (typically, $N_u^m \gg N_l^m$), $\bm{x}^n \in \mathcal{X} \subseteq \mathbb{R}^d$ is the input sampled from a $d$-dimensional space, $\bm{y}^n \in \mathcal{Y} \subseteq {\{0,1\}}^K$ is the one-hot label. For clarity, we sometimes omit the superscript denoting the client index in the following contents.

With a slight abuse of notation, we denote $N_l^k$ and $N_u^k$ as the numbers of samples in class $k$ under $\mathcal{D}_l$ and $\mathcal{D}_u$ for an arbitrary client, i.e., $\sum_{k=1}^{K} N_l^k = N_l$ and $\sum_{k=1}^{K} N_u^k = N_u$. In this paper, we assume that both $\mathcal{D}_l$ and $\mathcal{D}_u$ exhibit class imbalance, that is, $\exists i,j \in \{1,2,\dots, K\}$ for which the ratio $\frac{N_l^i}{N_l^j}$ is significantly greater than $1$. 
In other words, the label prior distribution $\{p_l^1, \dots, p_l^K\}$ shifts from a uniform distribution $\{\frac{1}{K}\}^K$.
This assumption is similarly applicable for $\mathcal{D}_u$.

Furthermore, we consider the setting that both intra-client and inter-client data heterogeneity exist in the FL system.
Intra-client heterogeneity refers to the varied distributions of labeled and unlabeled data within a single client, that is, $\forall m \in \{1,2,\dots,M\}, \mathcal{D}_u^m \neq \mathcal{D}_l^m$.  
Inter-client heterogeneity, on the other hand, pertains to the dissimilar mixture distributions of both labeled and unlabeled data across clients, i.e., $\forall i, j \in \{ 1,2,\cdots M\}, i \neq j, \text{it holds that} \ \mathcal{D}_l^i + \mathcal{D}_u^i \neq \mathcal{D}_l^j + \mathcal{D}_u^j$.

The final objective of FSSL is to learn a global model 
$f(\bm{x};\bm{w}): \mathcal{X} \rightarrow \mathcal{Y}$ parameterized by $\bm{w}$ that can generalize well to a balanced test dataset whose label prior distribution is $\{\frac{1}{K}\}^K$.
Given the input $\bm{x}^n$, we denote its corresponding output logits as $\bm{z}(\bm{x}^n):= f(\bm{x}^n;\bm{w})$, and the normalized prediction probability after softmax layer as $\bm{p}(\bm{y}|\bm{x}^n) := \sigma(f(\bm{y}|\bm{x}^n;\bm{w}))$, where $\sigma(\cdot)$ is the softmax function.
The detailed framework of FSSL is explained as follows.

\subsection{Framework of FSSL}
During each global round, the server first selects a random subset of clients $\mathcal{S}$ based on the activation rate $C$ and broadcasts the global model $\bm{w}$ to these clients. Subsequently, these clients perform local training for $E$ epochs using $\bm{w}$ as initial weights, resulting in the updated local model $\bm{w}_m$. Finally, the selected clients upload their local models $\bm{w}_m$ to the server for model aggregation. The training paradigms of labeled and unlabeled data in local clients are as follows.  

For labeled data, the standard cross-entropy loss is applied to the weakly augmented version of samples to promote the discriminative objective, as shown below:
\begin{equation}
    \mathcal{L}_s =  \frac{1}{N_l} \sum_{n=1}^{N_l} \mathrm{H}(\bm{y}^n, \bm{p}(\bm{y}|\alpha(\bm{x}^n))),
\end{equation}
where $\alpha(\cdot)$ is the weak augmentation function, $\bm{p}(\bm{y}|\alpha(\bm{x}^n))$ is the prediction probability for $\alpha(\bm{x}^n)$, and $\mathrm{H}(\bm{p}_1, \bm{p}_2)$ is entropy between probability distributions $\bm{p}_1$ and $\bm{p}_2$.

For unlabeled data, the samples are pseudo-labeled using the trained model, after which they are incorporated into the training process.
Specifically, for a given unlabeled sample $\bm{x}^n$, the model first generates the probability on its weakly augmented version. Then the pseudo-label is calculated by:
\begin{equation}
    \hat{\bm{y}}^n = \arg \max (\bm{p}(\bm{y}|\alpha(\bm{x}^n))),
\end{equation}
where $\arg \max(\cdot)$ is the function that converts a probability distribution into a one-hot label based on its maximum value.

To enhance the model generalization, the consistency loss is applied to unlabeled data by minimizing the entropy between the pseudo-label and the prediction of its strong augmented version.

During the training, only those unlabeled samples that exhibit high confidence are selected to participate in further training.
Consequently, the overall optimization objective for the unlabeled data can be expressed as follows:
\begin{align}
    \mathcal{L}_u = \frac{1}{N_u} \sum_{n=1}^{N_u} & \mathbb{1} (\max (\bm{p}(\bm{y}|\alpha(\bm{x}^n))) \ge \tau) \cdot \\
    & \mathrm{H}(\hat{\bm{y}}^n, \bm{p}(\bm{y}|\mathcal{A}(\bm{x}^n))),
\end{align}
where $\tau$ is the threshold, $\mathbb{1}(\cdot)$ is the indicator function, $\mathcal{A}(\cdot)$ is the strong augmentation function.

The overall optimization objective of local training on clients is expressed as:
\begin{equation}
    \mathcal{L}=\mathcal{L}_s+\lambda \mathcal{L}_u,
\end{equation}
where $\lambda$ is used to balance these two loss terms. 

After local training, the selected clients send the latest local models to the server for model aggregation, as shown below:
\begin{equation}
    \bm{w}^{t+1} =\frac{1}{\left | \mathcal{S} _t \right | } \sum_{m \in \mathcal{S} _t} \bm{\beta}_m \cdot \bm{w}_m^t ,
\end{equation}
where $\left | \mathcal{S} _t \right |$ is the number of selected clients in round $t$, $\bm{w}^{t}_m$ is the local model on client $m$ in round $t$, $\bm{\beta}_m$ is the aggregate weight for $\bm{w}^{t}_m$, $\bm{w}^{t+1}$ is the global model in round $t+1$.

\section{FedDB: Detaching Prior Bias in FSSL}
This section details the framework of FedDB and its two key techniques: debiased pseudo-labeling (DPL) and debiased model aggregation (DMA).

\subsection{Framework Overview of FedDB}
\begin{figure}[!htp]
\centering
\includegraphics[width=3.3in]{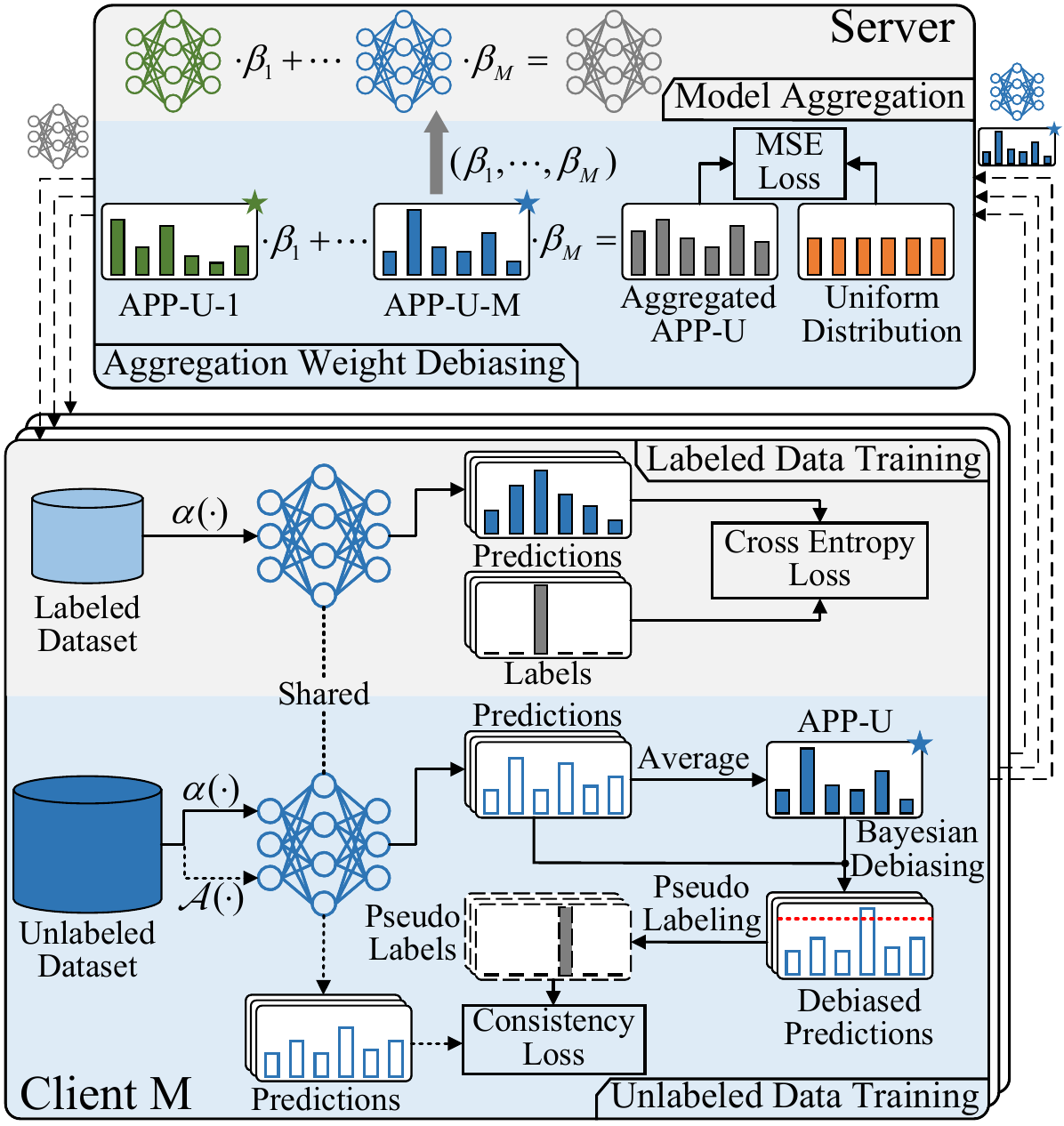}
\caption{Framework overview of FedDB.}
\label{framework}
\end{figure}

Figure~\ref{framework} illustrates the framework of FedDB. During the training, each global round consists of the following steps:
\begin{enumerate}[(1)]
    \item The server selects a subset of clients for training and broadcasts the global model to these clients;
   \item The clients perform inference on unlabeled data and calculate APP-U to estimate the prior bias;
   \item The clients perform DPL on unlabeled data using APP-U; 
   \item  The clients train the model utilizing both labeled data and pseudo-labeled data;
   \item The clients upload local models and APP-U to the server. The server performs DMA using APP-U from clients;
   \item Repeating steps 1-5 until the global model converges.
\end{enumerate}

\subsection{Prior Bias Estimation}
In this paper, we consider an FSSL setting where both the labeled data and unlabeled data are class imbalanced.
In such a case, the model's predictions can skew towards certain classes, owing to the biased label prior in the training data.
This skew contradicts the training objective of FSSL, which is to achieve uniform performance across all classes.

To investigate the impact of class imbalance on model training in FSSL, we conduct preliminary experiments using the CIFAR10 dataset.
We establish a scenario with 10 clients, each participating in model training in every round. 
The number of labeled and unlabeled samples is set to $4000$ and $46000$, respectively.
The class imbalance is created by the Dirichlet distribution, as declared in Section \ref{Experiments}.

As shown in Figure \ref{class-wise-accuracy}, both the local and global models exhibit a biased prediction towards certain classes.
However, estimating the above bias in FSSL is challenging due to the data heterogeneity and imprecision in pseudo-labeling.
By extensive experiments, we discover that the prior bias can be effectively approximated by the Average Prediction Probability on Unlabeled Data (APP-U). 
Specifically, for client $m$, the APP-U, denoted by $\overline{\bm{p}}_m$, can be calculated by:
\begin{equation}
    \overline{\bm{p}}_m=\frac{ {\textstyle \sum_{n=1}^{N_u^m}} \bm{p}(\bm{y}|\alpha(\bm{x}_{u}^{n})) }{N_u^m} ,
    \label{equ:estimate}
\end{equation}
where $N_u^m$ denotes the total number of unlabeled samples on client $m$, $\bm{p}(\bm{y}|\alpha(\bm{x}_{u}^{n}))$ is the prediction probability of the weak augmentation of sample $\bm{x}_{u}^n$.

We adopt JS divergence as a metric to quantify the disparity between two distributions.
A larger JS divergence indicates a greater disparity between the distributions.
Taking the class-wise accuracy on a balanced test dataset as the ground truth bias, we calculate the JS divergence between it and either APP-U or the labeled data distribution. As shown in Figure \ref{JS_Divergence}, for both local and global models, the JS divergence between APP-U and the ground truth is significantly smaller than that between the labeled data distribution and the ground truth. This demonstrates the effectiveness of APP-U as a metric for quantifying prior bias in FSSL.

\subsection{Debiased Pseudo-Labeling}
In this subsection, we detail the procedure of DPL.
Given an FL model parameterized by $\bm{w}$, we first obtain the prediction probability $\bm{p}_s(\bm{y}|\bm{x})$ by applying a softmax function to unnormalized logits, as illustrated below:
\begin{equation}
    \bm{p}_s(y|\bm{x}) = \frac{e^{\bm{z}(\bm{x})[y]}}{ {\textstyle \sum_{k=1}^{K}}e^{\bm{z}(\bm{x})[k]}} ,
    \label{softmax}
\end{equation}
where $\bm{z}(\bm{x})[y]$ is the $y$-th unnormalized logit. 

By applying the Bayes' theorem to $\bm{p}_s(y|\bm{x})$, we obtain:
\begin{equation}
\bm{p}_s(y|\bm{x}) = \frac{\bm{p}_s(y)\bm{p}_s(\bm{x}|y)}{ {\textstyle \sum_{k=1}^{K}} \bm{p}_s(k)\bm{p}_s(\bm{x}|k)}.
\label{Bayesian}
\end{equation}

Due to the class imbalance in our FSSL settings, the prior distribution $\bm{p}_s(k)$, as outputted by the model, is biased towards certain majority classes. This leads to a biased prediction probability $\bm{p}_s(y|\bm{x})$, causing the model to be overconfident in these majority classes. 
The objective of DPL is to seek a conditional probability $\bm{p}_t(y|\bm{x})$ that is robust across all classes, given the estimation of the model's biased prior $\overline{\bm{p}}$, as defined in Eq. (\ref{equ:estimate}). 

Following previous studies \cite{tian2020posterior,kairouz2021advances,hong2021disentangling}, we assume that the class conditional likelihoods are the same in both the biased and debiased predictions, i.e., $\bm{p}_t(\bm{x}|y) = \bm{p}_s(\bm{x}|y)$.
By rearranging Eq. (\ref{softmax}) and Eq. (\ref{Bayesian}), we have:
\begin{equation}
    \begin{aligned}
    \ln(\bm{p}_t(y)\bm{p}_t(\bm{x}|y)) = & \bm{z}(\bm{x})[y]+\ln(\bm{p}_t(y))-\ln(\bm{p}_s(y)) \\
    &+ \ln({\textstyle \sum_{k=1}^{K}} \bm{p}_s(k)\bm{p}_s(\bm{x}|k)) \\
    &- \ln( {\textstyle \sum_{k=1}^{K}} e^{\bm{z}(\bm{x})[k]}). 
\end{aligned}
\end{equation}

Recalling that:
\begin{equation}
    \bm{z}(\bm{x})[y] = \ln \bm{p}_s(y|\bm{x}) + \ln {{\textstyle \sum_{k=1}^{K}} \bm{p}_s(k)\bm{p}_s(\bm{x}|k)}.
\end{equation}

We derive the following debiased posterior probability:
\begin{equation}
    \begin{aligned}
    \bm{p}_t(y|\bm{x}) 
    &= \frac{\bm{p}_t(y)\bm{p}_t(\bm{x}|y)}{\sum_{k=1}^{K} \bm{p}_t(k)\bm{p}_t(\bm{x}|k)} \\
    &= \frac{\bm{p}_s(y|\bm{x})\bm{p}_t(y)/\bm{p}_s(y) }{ {\textstyle \sum_{k=1}^{K}} \bm{p}_s(k|\bm{x})\bm{p}_t(k)/\bm{p}_s(k)},
\end{aligned}
\end{equation}

where $\bm{p}_t(k)$ is a uniform distribution that is robust for all classes. 
By applying the estimated bias $\overline{\bm{p}}$ as the approximation of the prior bias $\bm{p}_s$,
we can obtain the debiased prediction probability of unlabeled data as follows:
\begin{equation}
    \hat{\bm{p}} = \frac{\bm{p}(\bm{y}|\bm{x})/\overline{\bm{p}}}{ {\textstyle \sum_{k = 1}^{K}} \bm{p}(k|\bm{x})/\overline{\bm{p}}_k}.
    \label{debiased}
\end{equation}

Intuitively, Eq. (\ref{debiased}) serves as a regularization term that smooths the prediction probabilities of the majority classes and sharpens these of the minority classes, which can alleviate the prior bias introduced by the heterogeneous data. 
The detailed procedures of DPL are shown in Algorithm \ref{algorithm: DPL}.

\begin{algorithm}[t]
    \caption{DPL: Debiased Pseudo-labeling}\label{algorithm: DPL}
    \begin{small}
    \BlankLine
    \KwIn{Confidence threshold $\tau$}
    \KwOut{Debiased pseudo-labels $\hat{\bm{Y}}$, APP-U $\overline{\bm{p}}$}
    \DontPrintSemicolon
    
    $\overline{\bm{p}}=\frac{ {\textstyle \sum_{n=1}^{N_u}} \bm{p}(\bm{y}|\alpha(\bm{x}_{u}^n)) }{N_u}$, 
    $\hat{\bm{Y}} := \{\}$\;
    \For {$n=1,2,...,N_u$}{
    $\hat{\bm{p}}^n := \frac{\bm{p}(\bm{y}|\alpha(\bm{x}_u^n))/\overline{\bm{p}}}{{\textstyle \sum_{k=1}^{K}} \bm{p}(k|\alpha(\bm{x}_u^n))/\overline{\bm{p}}_{k} }$ \;
    \If {$\max(\hat{\bm{p}}^n) \ge \tau$}{
    $\hat{\bm{Y}} := \hat{\bm{Y}} \oplus \arg \max (\hat{\bm{p}}^n$) \;
    }
    \Else {
    $\hat{\bm{Y}} := \hat{\bm{Y}} \oplus \{0\}^{K}$}
    }
    Return $\hat{\bm{Y}}$, $\overline{\bm{p}}$ \;
    \end{small}
\end{algorithm}

\subsection{Debiased Model Aggregation}
The objective of DMA is to computing aggregation weights that enable the model to perform uniformly across all classes.
During each local updating round, the activated clients send their accumulated APP-U $\overline{\bm{p}}_m$ and their latest models $\bm{w}_m$ to the server. Then we can get the aggregated APP-U as follows:
\begin{equation}
    \overline{\bm{p}}_{aggr} ={\textstyle \sum_{m \in \mathcal{S}_t}} \bm{\beta}_m \overline{\bm{p}}_m,
\end{equation} 
where $\bm{\beta}_m$ denotes the aggregation weight for client $m$. To achieve a more balanced model, we expect $\overline{\bm{p}}_{aggr}$ to be more uniform, leading to the following optimization objective:

\begin{equation}
\begin{aligned}
    &\min_{\bm{\beta}} \mathcal{L} _{aggr} = \sqrt{ \textstyle \sum_{m=1}^M  (\overline{\bm{p}}_{aggr} - \bm{p}_{t})^2} \\ &
    \text{s.t.} \textstyle \sum_{m \in \mathcal{S}_t} \bm{\beta}_m = 1, 
\end{aligned}
\end{equation}
where $\bm{p}_{t} = \{\frac{1}{K}\}^K$ is the uniform distribution over $K$ classes, identical to the test dataset. 
In FedDB, we utilize the gradient descent algorithm to solve the above optimization problem.

After obtaining the aggregation weights $\bm{\beta}$, we aggregate client models and update the global model as follows:
\begin{equation}
    \bm{w}^{t+1} = \textstyle \sum_{m \in \mathcal{S}_t}\bm{\beta}_m \cdot \bm{w}_m^t ,
\end{equation}
where $\bm{w}^t_m$ is the local model of client $m$ at last round, $\bm{w}^{t+1}$ is the global model. $\bm{w}^{t+1}$ is then broadcast to the activated client for further updates. 
The processes of DMA and FedDB are presented in Algorithms \ref{algorithm: DMA} and \ref{algorithm: FedDB}, respectively.

\begin{algorithm}[t]
    \caption{DMA: Debiased Model Aggregation}\label{algorithm: DMA}
    \begin{small}
    \BlankLine
    \KwIn{Local models$\{\bm{w}_m\}_{m=1}^{M}$, local APP-U $\{\overline{\bm{p}}_m\}_{m=1}^{M}$, updating epochs $E_{aggr}$, learning rate $\eta_{aggr}$}
    \KwOut{Global weight $\bm{w}$}
    \SetKwFunction{Serverexe}{Server executes}
    \SetKwFunction{Clientupd}{ClientUpdate}
    \SetFuncSty{textbf}
    \DontPrintSemicolon
    Initialize $\bm{\beta}$ as $\{\frac{1}{M}\}^M$
    
    \For{$e=1,2,...,E_{aggr}$}{
        $\overline{\bm{p}}_{aggr} \gets  {\textstyle \sum_{m=1}^{M} \bm{\beta}_m \overline{\bm{p}}_m} $ \;
        $\mathcal{L} _{aggr} = \sqrt{ \textstyle \sum_{m=1}^M  (\overline{\bm{p}}_{aggr} - \bm{p}_{t})^2}$ \;
        $\bm{\beta} \gets \bm{\beta} - \eta_{aggr} \nabla \mathcal{L}_{aggr} $\;
        $\bm{\beta} = \sigma(\bm{\beta})$ \;
    }
    $\bm{w} \gets  {\textstyle \sum_{m=1}^{M} \bm{\beta}_m \bm{w}_m} $\;
    
    Return $\bm{w}$\;
    \end{small}
\end{algorithm}

\begin{algorithm}[!ht]
    \caption{FedDB: Detaching Prior Bias in FSSL}\label{algorithm: FedDB}
    \begin{small}
    \BlankLine
    \KwIn{Client number $M$, client activate rate $C$, global rounds $T$, update epochs $E$ and $E_{aggr}$, learning rate $\eta$ and $\eta_{aggr}$, threshold $\tau$, unlabeled loss weight $\lambda$, momentum accumulation coefficient $\gamma$}
    \KwOut{Global model $w^T$}
    \SetKwFunction{Serverexe}{Server executes}
    \SetKwFunction{Clientupd}{ClientUpdate}
    \SetFuncSty{textbf}
    \DontPrintSemicolon
    
    \Serverexe:\;
    Initialize $\bm{w}^0$ \;
    \For{$t=1,2,...,T$}{
        $\mathcal{S}_t \gets$ randomly select $M \cdot C$ clients\;
        \For{each client in $m \in S_t$ \textbf{in parallel}}{
            %send the global model $w^t$ to client $m$ \;
            $\bm{w}_m^t$, $\overline{\bm{p}}_m^t$ $\gets$ \textbf{ClientUpdate}$(\bm{w}^{t-1})$ \;
        }
        $\bm{w}^{t} \gets$ \textbf{DMA}($\{\bm{w}_m^t\}_{m \in \mathcal{S}_t}$, $\{\overline{\bm{p}}_m^t\}_{m \in \mathcal{S}_t}$, $E_{aggr}$, $\eta_{aggr}$)
    }
    Return $\bm{w}^T$\;
    \Clientupd{$\bm{w}^t$}\;
    $\hat{\bm{Y}}, \overline{\bm{p}} \gets$ \textbf{DPL}($\tau$)\;
    \For{$e=1,2,...,E$}{
        $\overline{\bm{p}}^e=\frac{ {\textstyle \sum_{n=1}^{N_u}} \bm{p}(\bm{y}|\alpha(\bm{x}_{u}^n)) }{N_u}$\;
        $\mathcal{L}_s =  \frac{1}{N_l} \sum_{n=1}^{N_l} \mathrm{H}(\bm{y}^n, p(\bm{y}|\alpha(\bm{x}_l^n)))$\;
        \setlength{\abovedisplayskip}{-1pt}
        \setlength{\belowdisplayskip}{-10pt}
        \hspace*{-0.1\leftmargin} \parbox{\textwidth}{
        \begin{flalign*}
            \mathcal{L}_u = \textstyle \frac{1}{N_u} \textstyle \sum_{n=1}^{N_u} \mathbb{1} & (\max(\hat{\bm{Y}}^n)) \ge \tau) \cdot & \\ 
            &\mathrm{H}(\hat{\bm{Y}}^n, p(\bm{y}|\mathcal{A} (\bm{x}_u^n)))&
        \end{flalign*}} \;
        
        $\mathcal{L} \gets \mathcal{L}_s+\lambda \mathcal{L}_u$\;
        
        $\bm{w}^e \gets \bm{w}^{e-1} - \eta \nabla \mathcal{L}$; \ $\overline{\bm{p}} \gets \gamma \overline{\bm{p}} + (1-\gamma) \overline{\bm{p}}^e $ \;
    }
    Return $\bm{w}^E$, $\overline{\bm{p}}$
    
    \end{small}
\end{algorithm}
\vspace{-7pt}

\section{Experiments} \label{Experiments}
This section details the experimental results in various settings to demonstrate the effectiveness of FedDB. 

\subsection{Experimental Setup}
\paragraph{Datasets.}
We evaluate FedDB on three benchmark datasets, including CIFAR10, SVHN, and CIFAR100.
Initially, a balanced labeled dataset is separated from the original training dataset, with the residual data designated as the unlabeled dataset.
When distributing these training data to clients, we sample data from a Dirichlet distribution $\bm{q} \sim \text{Dir}(\delta \bm{p})$, where $\bm{p}$ is the class-wise prior distribution and $\delta$ is a parameter that modulates the heterogeneity among clients. 
A higher value of $\delta$ correlates with reduced data heterogeneity.
To enrich the unlabeled dataset, we add the samples from the labeled dataset to the unlabeled dataset after discarding their labels.
We conduct experiments in IID setting and Non-IID settings with $\delta =\{0.1,0.3\}$.
In the IID setting, the total number of labeled samples is set to $4000, 1000, 10000$ for CIFAR10, SVHN and CIFAR100, respectively. For Non-IID setting, the total number of labeled data is set to $4000$ for CIFAR10 and SVHN,  and $10000$ for CIFAR100.
The test dataset from the original dataset is used for model evaluation. 

\paragraph{Benchmark Methods.}
We compare FedDB against the following benchmark methods:
\begin{itemize}
    \item \textbf{FedAvg} \cite{FedAvg}: The FedAvg method is applied in a constrained scenario where each client utilizes only the small labeled dataset for training.
    \item \textbf{FixMatch} \cite{Fixmatch}: This method is a basic adaptation of FixMatch within FedAvg framework. 
    \item \textbf{FedMatch} \cite{FedMatch}: FedMatch introduces the inter-client consistency loss to maximize the agreement between local models. 
    \item \textbf{FedRGD} \cite{FedRGD}: It mitigates the model bias by reducing gradient divergence among clients. 
    \item \textbf{SemiFL} \cite{SemiFL}: SemiFL adopts alternate training between server and clients. Here, we adopts its client-side training due to the lack of training samples on the server in our scenario.  
    \item \textbf{Methods combining DPL.} We also conduct experiments that integrate DPL with benchmark methods. These hybrid methods are denoted as \textbf{Method-FedDPL}.
\end{itemize}

\paragraph{Implementation Details.}
We primarily follow the experimental settings adopted in prior works of FSSL \cite{FedMatch}. 
There are a total of $100$ clients participating in the training, with $10$ active clients $(C=0.1)$ engaged in each global round. 
The local training epoch is set to $E = 5$ and the epoch for updating the model aggregation weights is set to $E_{aggr} = 100$. 
All experiments are executed for $800$ global rounds. 
We employ Wide ResNet28x2 in our experiments.
The SGD optimizer is adopted for model training, operating at learning rates $\eta=0.03$ for local updating and $\eta_{aggr}=1.0$ for aggregation, complemented by a momentum of $0.9$. 
Due to the limited number of samples on clients, we feed all training data simultaneously to the model during local training. 
The confidence threshold for pseudo-labeling is set to $\tau=0.95$. 
The data augmentation operation is consistent with those described in FixMatch \cite{Fixmatch}. 
All experiments are repeated for $4$ times and we report the mean and standard deviation of the best accuracy during training.

\subsection{Results on Benchmark Datasets}

\begin{figure}[!ht]
    \centering
    \includegraphics[width=3.3in]{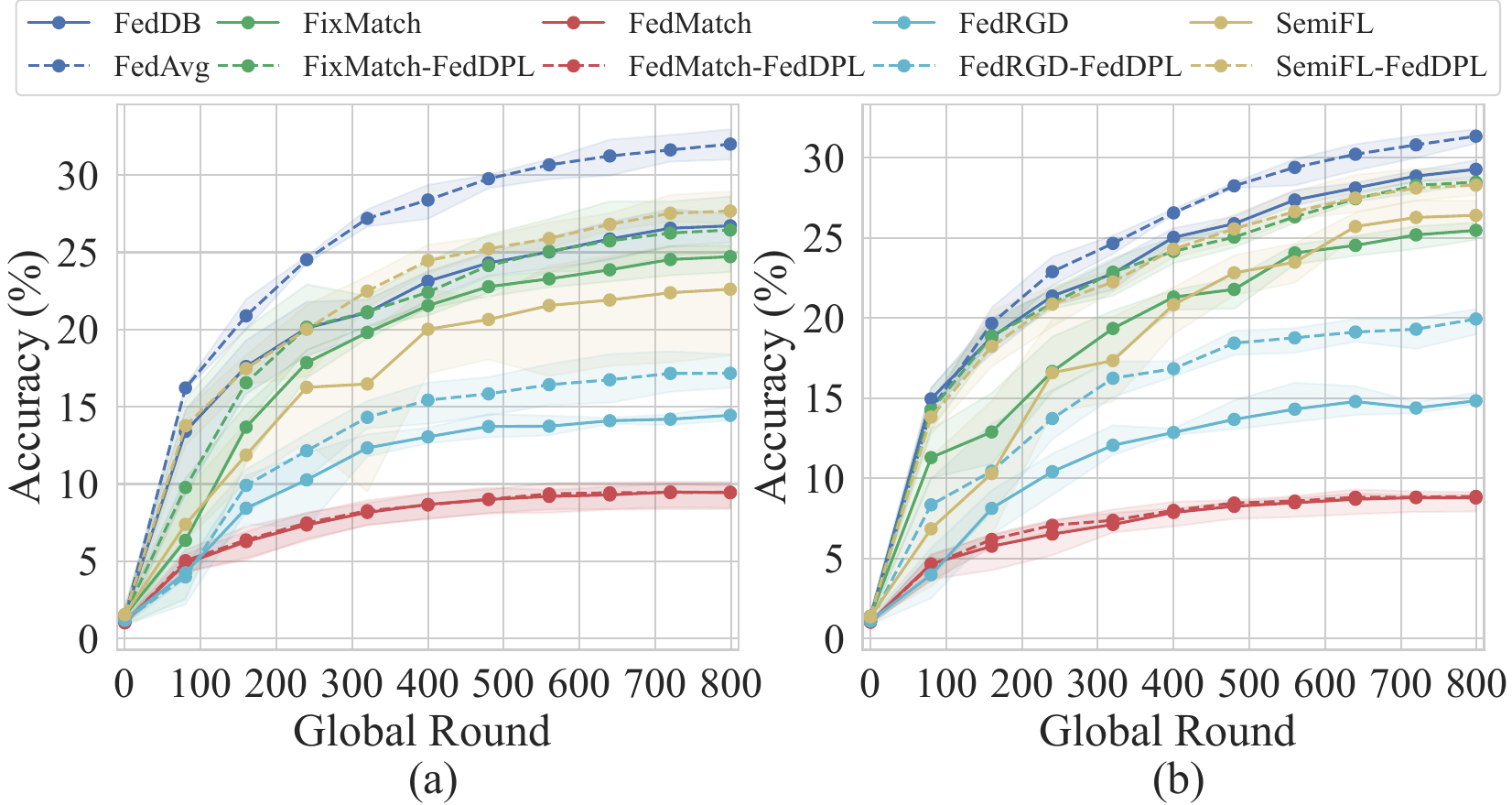}
\caption{Convergence curve on CIFAR100. (a) IID, (b)Non-IID with $\delta=0.3$.}
\label{fig: convergence}
\end{figure}

The experimental results are presented in Tables \ref{tab: iid} - \ref{tab: alpha0.1}, where values inside the parentheses represent the mean, and values outside the parentheses represent the standard deviation of multiple experiments.
It can be observed that with the same number of labeled samples, the accuracy of all methods decreases as $\delta$ decreases, demonstrating that data heterogeneity is a key factor harming model performance.
FedAvg, despite its simplicity, serves as a reliable benchmark method, particularly as the dataset difficulty increases (e.g., CIFAR100). 
This issue is also noted by \cite{SemiFL}.
This demonstrates that improperly incorporating unlabeled data into training can negatively impact the model's training.
Compared with other FSSL methods, FedDB enhances test accuracy, demonstrating the effectiveness of FedDB in the FSSL scenario.
The same conclusion can also be drawn from Figure \ref{fig: convergence}.

\begin{table}[!ht]
\centering
\resizebox{3.4in}{!}{
\begin{tabular}{cccc}
\toprule
Dataset & CIFAR10 & SVHN & CIFAR100 \\ \midrule
FedAvg & 58.42(0.61) & 25.10(0.76) & 32.00(0.80) \\ \midrule 
FixMatch & 65.80(2.72) & 87.44(1.35) & 24.72(0.73) \\
FedMatch & 39.63(1.66) & 25.09(5.40) & 9.44(0.66) \\
FedRGD & 63.27(1.47) & 81.04(2.43) & 14.45(0.42) \\
SemiFL & 57.24(7.96) & 85.58(10.03) & 22.61(3.07) \\ \midrule 
FixMatch-FedDPL & 66.97(2.84) & 88.00(0.67) & 26.44(1.73) \\
FedMatch-FedDPL & 43.06(3.16) & 25.90(3.12) & 9.47(0.79) \\
FedRGD-FedDPL & 64.75(1.20) & 81.24(5.36) & 17.17(0.98) \\
SemiFL-FedDPL & 68.46(3.61) & 86.77(1.79) & 27.67(0.89) \\ \midrule
FedDB & 67.32(2.31) & 86.75(0.90) & 26.71(0.87) \\ \bottomrule
\end{tabular}}
\caption{Experimental results in the IID setting.}
\label{tab: iid}
\end{table}

\begin{table}[!ht]
\centering
\resizebox{3.4in}{!}{
\begin{tabular}{cccc}
\toprule
Dataset & CIFAR10 & SVHN & CIFAR100 \\ \midrule
FedAvg & 47.72(1.95) & 69.44(6.21) & 31.34(0.36) \\ \midrule 
FixMatch & 50.99(2.49) & 86.61(0.19) & 25.47(0.46) \\
FedMatch & 38.64(2.49) & 26.04(4.85) & 8.77(0.57) \\
FedRGD & 51.45(2.39) & 86.89(3.21) & 14.83(0.34) \\
SemiFL & 50.07(1.05) & 76.11(6.3) & 26.40(0.81) \\ \midrule 
FixMatch-FedDPL & 53.92(3.41) & 85.87(0.51) & 28.47(0.13) \\
FedMatch-FedDPL & 39.17(2.10) & 27.02(3.13) & 8.87(0.11) \\
FedRGD-FedDPL & 51.57(1.67) & 87.00(1.31) & 19.94(0.75) \\
SemiFL-FedDPL & 55.42(2.57) & 87.61(0.91) & 28.29(0.73) \\ \midrule
FedDB & 55.00(1.17) & 85.99(0.49) & 29.28(0.51) \\ \bottomrule
\end{tabular}}
\caption{Experimental results in the Non-IID setting with $\delta=0.3$.}
\label{tab: alpha0.3}
\end{table}

\subsection{Effectiveness of DPL}
\begin{figure}[!ht]
    \centering
    \includegraphics[width=3.3in]{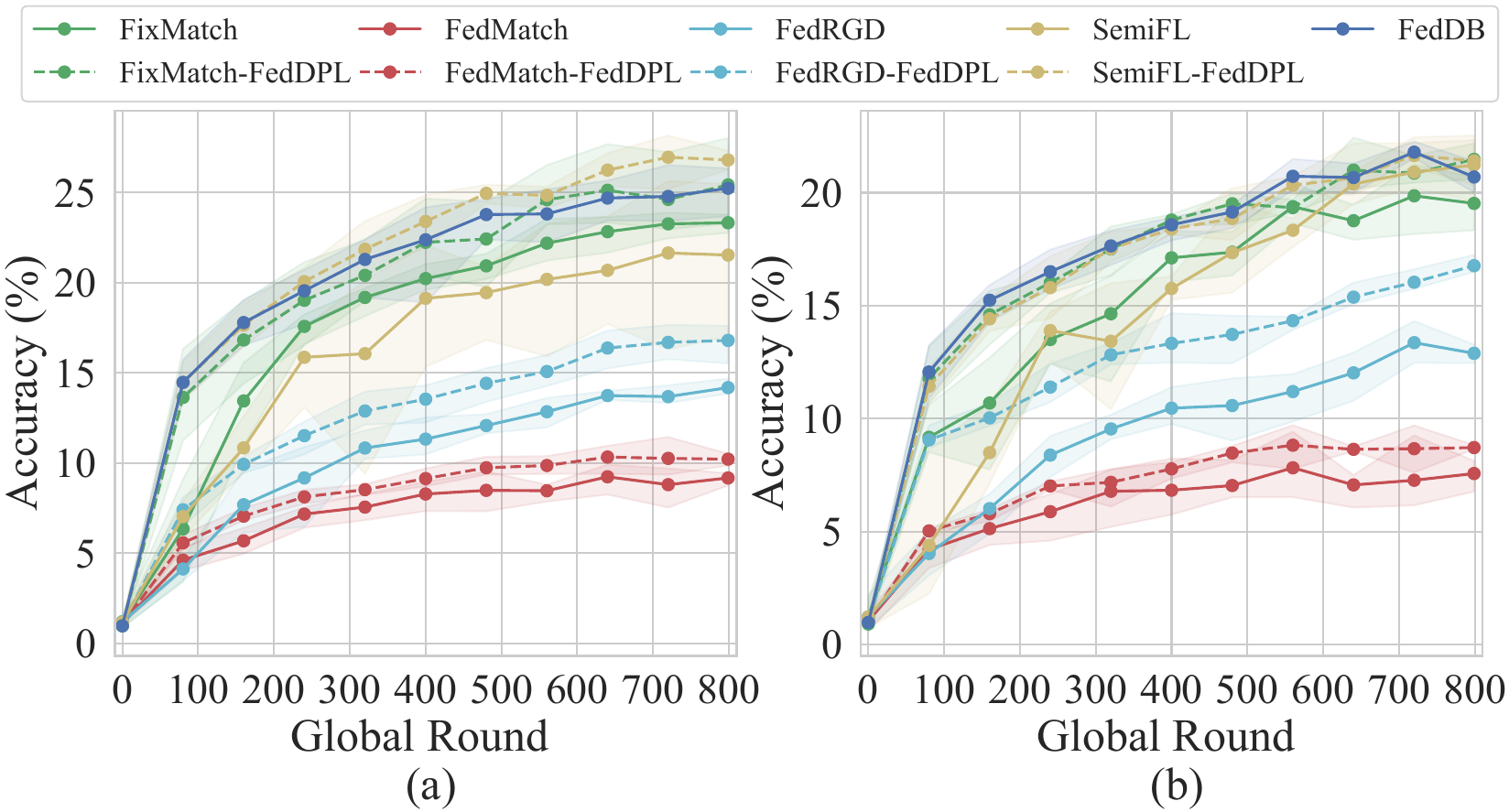}
\caption{Accuracy of pseudo labels on CIFAR100. (a) IID, (b)Non-IID with $\delta=0.3$.}
\label{fig: pseudo_acc}
\end{figure}

\begin{figure}[!ht]
    \centering
    \includegraphics[width=3.3in]{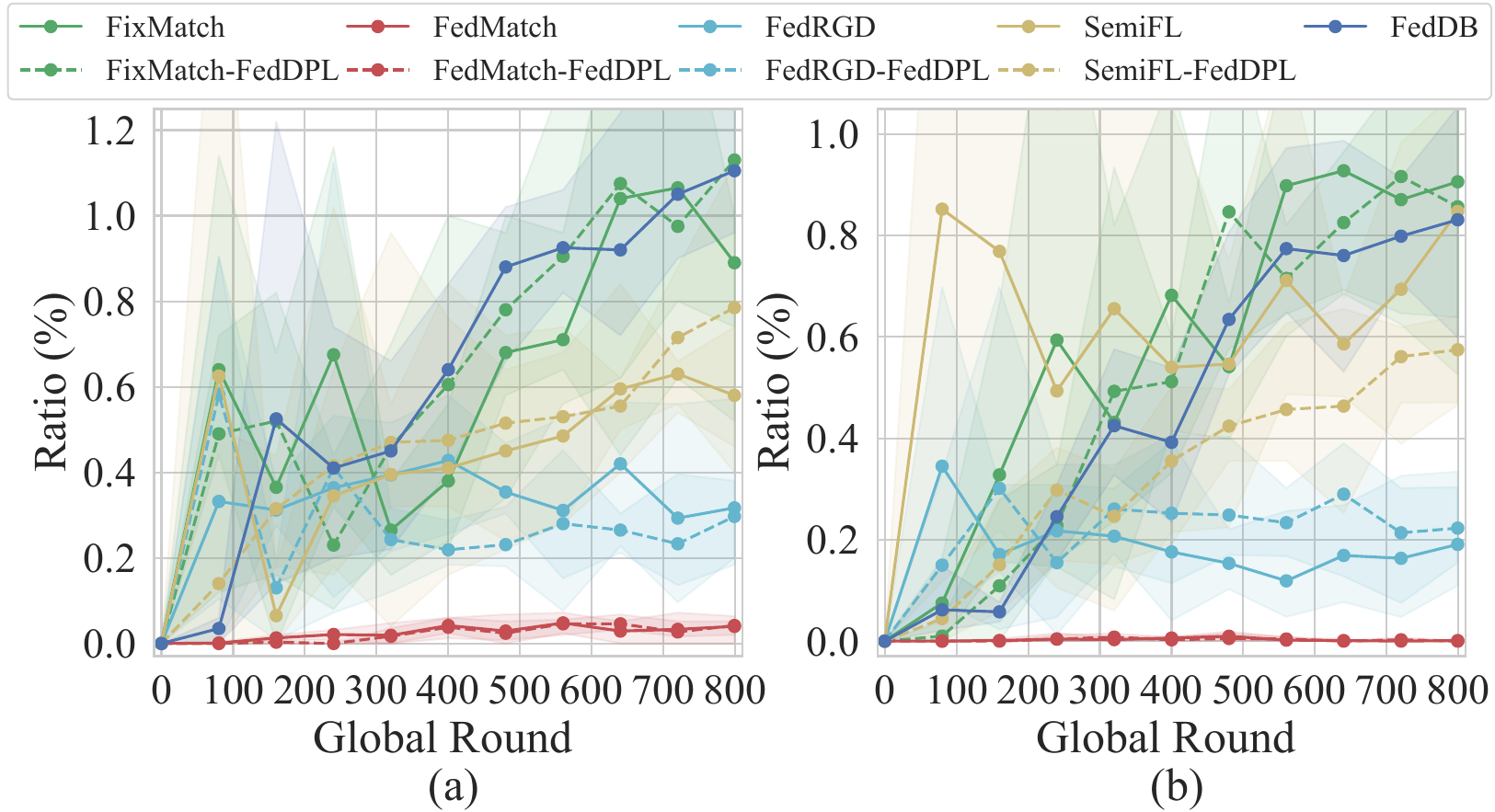}
\caption{Ratio of unlabeled samples that are finally assigned with pseudo-labels on CIFAR100. (a) IID, (b)Non-IID with $\delta=0.3$.}
\label{fig: pseudo_ratio}
\end{figure}

As illustrated in Table \ref{tab: ablation}, employing DPL results in substantial gains for FedDB.
Furthermore, DPL can be regarded as a convenient plug-in that can be easily integrated into existing FSSL methods utilizing pseudo-labeling. 
As shown in Tables \ref{tab: iid} - \ref{tab: alpha0.1}, introducing DPL to existing FSSL methods effectively enhances their performance.
Figure \ref{fig: pseudo_acc} displays the accuracy of pseudo-labels during training. It indicates that DPL effectively enhances the accuracy of these pseudo-labels, which in turn benefits FSSL training.
Figure \ref{fig: pseudo_ratio} presents the ratio of pseudo-labeled samples in the unlabeled data. 
However, introducing DPL does not consistently improve the ratio of pseudo-labeled samples, as the model in FSSL is challenging to train, making it difficult for samples to be pseudo-labeled.

\begin{table}[t]
\centering
\resizebox{3.4in}{!}{
\begin{tabular}{cccc}
\toprule
Dataset & CIFAR10 & SVHN & CIFAR100 \\ \midrule
FedAvg & 33.53(1.9) & 32.21(1.52) & 28.78(0.53) \\ \midrule 
FixMatch & 35.14(1.53) & 74.31(2.07) & 25.90(1.06) \\
FedMatch & 31.12(2.69) & 12.66(3.34) & 7.50(0.99) \\
FedRGD & 35.33(3.73) & 38.20(5.64) & 18.04(1.59) \\
SemiFL & 33.72(1.87) & 72.76(6.19) & 25.82(0.44) \\ \midrule 
FixMatch-FedDPL & 37.13(3.22) & 76.29(1.00) & 27.76(0.85) \\
FedMatch-FedDPL & 32.26(2.75) & 16.94(1.28) & 7.66(0.43) \\
FedRGD-FedDPL & 35.59(3.49) & 38.76(2.67) & 18.98(0.58) \\
SemiFL-FedDPL & 37.84(2.33) & 74.54(7.51) & 27.62(1.00) \\ \midrule
FedDB & 37.95(2.21) & 76.20(1.31) & 27.99(1.28) \\ \bottomrule
\end{tabular}}
\caption{Experimental results in the Non-IID setting with $\delta=0.1$.}
\label{tab: alpha0.1}
\end{table}

\subsection{Effectiveness of DMA}
As shown in Table \ref{tab: ablation}, DMA generally contributes positively to FedDB in most scenarios. However, its impact differs among various datasets.
More specifically, DMA consistently results in improved outcomes on the CIFAR10 and CIFAR100 datasets. 
Conversely, on the SVHN dataset, DMA can lead to performance decline in certain scenarios.
Upon detailed analysis, we ascribe this issue to the imbalanced distribution of the SVHN dataset, which contravenes the objective of FSSL that seeks for a balanced model.

\begin{table}[!ht]
\centering
\resizebox{3.4in}{!}{
\begin{tabular}{m{1cm}<{\centering}|cc|ccc}
\bottomrule
\multirow{4}{*}{IID} & DPL & DMA & CIFAR10 & SVHN & CIFAR100 \\ \cline{2-6}
& - & - & 65.80(2.72) & 87.44(1.35) & 24.72(0.73)  \\
& \checkmark & - & 66.97(2.84) & 88.00(0.67) & 26.44(1.73)  \\
& \checkmark & \checkmark & 67.32(2.31) & 86.75(0.90) & 26.71(0.87)  \\ \hline
\multirow{4}{*}{$\delta=0.3$} & DPL & DMA & CIFAR10 & SVHN & CIFAR100 \\ \cline{2-6}
& - & - & 50.99(2.49) & 86.61(0.19) & 25.47(0.46)  \\
& \checkmark & - & 53.92(3.41) & 85.87(0.51) & 28.47(0.13)  \\
& \checkmark & \checkmark & 55.00(1.17) & 85.99(0.49) & 29.28(0.51)  \\ \hline
\multirow{4}{*}{$\delta=0.1$} & DPL & DMA & CIFAR10 & SVHN & CIFAR100 \\ \cline{2-6}
& - & - & 35.14(1.53) & 74.31(2.07) & 25.90(1.06)  \\
& \checkmark & - & 37.13(3.22) & 76.29(1.00) & 27.76(0.85)  \\
& \checkmark & \checkmark & 37.95(2.21) & 76.20(1.31) & 27.99(1.28)  \\ \toprule
\end{tabular}}
\caption{Ablation studies on CIFAR10, SVHN, and CIFAR100.}
\label{tab: ablation}
\end{table}

\section{Conclusion}
In this paper, we propose FedDB to detach prior bias in FSSL with class imbalance.
At the local training level, FedDB debiases the pseudo-labeling using APP-U based on Bayes' theorem, encouraging a more balanced training data during the training.
At the global aggregation level, FedDB leverages APP-U across different clients to derive optimal aggregation weights, aiming to debias the global model.
Extensive experiments have shown the effectiveness of FedDB.

\section*{Acknowledgements}
This work was supported by the National Natural Science
Foundation of China under Grants 62372028 and 62372027.

\bibliographystyle{named}
\bibliography{reference}

\end{document}